%% file: root.tex
\documentclass[letterpaper, 10 pt, conference]{ieeeconf}  % Comment this line out if you need a4paper
\usepackage{enumerate}
\usepackage{amssymb}
\usepackage{amsmath}
\usepackage{textcomp}
\usepackage{xcolor}
\usepackage{multirow}
\usepackage{booktabs}
\usepackage{graphicx}
\usepackage{booktabs}  
\usepackage{threeparttable} 
\usepackage{hyperref}

\IEEEoverridecommandlockouts                              % This command is only needed if 
\title{\LARGE \bf
LoGoPlanner: Localization Grounded Navigation Policy with Metric-aware Visual Geometry
}

\author{Jiaqi Peng$^{*,1,2}$, Wenzhe Cai$^{*,2}$, Yuqiang Yang$^{*,2}$, Tai Wang$^{2, \dagger}$, Yuan Shen$^{1,2, \dagger}$ and Jiangmiao Pang$^{2}$ % <-this % stops a space
\thanks{$^{1}$Jiaqi Peng, Yuan Shen are with the Department of Electronic Engineering, Tsinghua University, $\dagger$Corresponding author.}%
\thanks{$^{2}$Jiaqi Peng, Wenzhe Cai, Yuqiang Yang, Tai Wang, Jiangmiao Pang and Yuan Shen are all with Shanghai AI laboratory, *Equal contribution.}%
}

\begin{document}

\maketitle
\thispagestyle{empty}
\pagestyle{empty}

\input{article/0.abstract}
\input{article/1.intro}
\input{article/2.related}

\input{article/3.setup}
\input{article/4.method}

\input{article/5.experiment}
\input{article/6.conclusion}
% \input{article/7.acknowledgement}


%%%%%%%%%%%%%%%%%%%%%%%%%%%%%%%%%%%%%%%%%%%%%%%%%%%%%%%%%%%%%%%%%%%%%%%%%%%%%%%%
\clearpage
\bibliographystyle{IEEEtran}
\bibliography{root}

\end{document}

%% file: article/0.abstract.tex
%%%%%%%%%%%%%%%%%%%%%%%%%%%%%%%%%%%%%%%%%%%%%%%%%%%%%%%%%%%%%%%%%%%%%%%%%%%%%%%%
\begin{abstract}
Trajectory planning in unstructured environments is a fundamental and challenging capability for mobile robots. Traditional modular pipelines suffer from latency and cascading errors across perception, localization, mapping, and planning modules. Recent end-to-end learning methods map raw visual observations directly to control signals or trajectories, promising greater performance and efficiency in open-world settings. However, most prior end-to-end approaches still rely on separate localization modules that depend on accurate sensor extrinsic calibration for self-state estimation, thereby limiting generalization across embodiments and environments.
% Path planning in unstructured environments is a fundamental yet challenging skill for robots. Traditional modular planning pipelines suffer from latency and compounding errors between modules. While recently end-to-end learning methods have been proposed, demonstrating high efficiency and ease of deployment in open-world scenarios. However, prior end-to-end approaches mainly replace perception, mapping, and planning modules, while still relying on additional localization modules for self-state updates. 
% This reliance undermines generalization across different embodiments, arbitrary viewpoints, and diverse environments. 
We introduce LoGoPlanner, a localization-grounded, end-to-end navigation framework that addresses these limitations by: (1) finetuning a long-horizon visual-geometry backbone to ground predictions with absolute metric scale, thereby providing implicit state estimation for accurate localization; (2) reconstructing surrounding scene geometry from historical observations to supply dense, fine-grained environmental awareness for reliable obstacle avoidance; and (3) conditioning the policy on implicit geometry bootstrapped by the aforementioned auxiliary tasks, thereby reducing error propagation.
% In this work, we propose LoGoPlanner, a localization-grounded end-to-end navigation framework that addresses these limitations through by: (1) using a long-temporal visual geometry backbone to estimate robot pose in a metric-aware manner, thereby grounding localization in metric scale and enabling implicit self-state estimation for planning; (2) reconstructing surrounding scene geometry using long-horizon obervation histories, providing fine-grained environmental awareness for reliable obstacle avoidance; and (3) conditioning the policy on implicit, feature-level predictions instead of raw predicted poses or pointclouds, thereby avoiding cascading errors while still benefiting from learned geometric priors. 
% We evaluate LoGoPlanner in both simulation and real-world settings. In simulation, our method achieves over a 10\% improvement compared to baselines that use oracle localization, demonstrating the effectiveness of implicit self-state estimation. In diverse real-world scenarios, LoGoPlanner generalizes across different robot platforms and environments, benefiting from its state consistency and metric-aware geometry perception.
We evaluate LoGoPlanner in both simulation and real-world settings, where its fully end-to-end design reduces cumulative error while metric-aware geometry memory enhances planning consistency and obstacle avoidance, leading to more than a 27.3\% improvement over oracle-localization baselines and strong generalization across embodiments and environments.
The code and models have been made publicly available on the \href{https://steinate.github.io/logoplanner.github.io/}{project page}.

\end{abstract}

%% file: article/1.intro.tex
%%%%%%%%%%%%%%%%%%%%%%%%%%%%%%%%%%%%%%%%%%%%%%%%%%%%%%%%%%%%%%%%%%%%%%%%%%%%%%%%
\section{INTRODUCTION}

Autonomous navigation, requiring robots to reliably reach specified goals in unstructured environments, remains a fundamental challenge in robotics. Traditional navigation pipelines are typically modular, decomposing the task into perception, localization, mapping, and planning~\cite{cao2022autonomous,fan2021step,wellhausen2021rough}. While this factorization improves interpretability and allows for component-level optimization, it often introduces compounding latency and suffers from cascading errors between modules~\cite{gog2021pylot}. These issues become particularly acute in real deployments, such as legged robots, where gait-induced vibrations in both cameras and IMUs reduce the accuracy of odometry and mapping—which in turn destabilizes downstream planning~\cite{wellhausen2021rough}.

End-to-end learning-based methods~\cite{roth2024viplanner,Yang-RSS-23,shah2023gnm,wijmans2019dd,cai2025navdp} have recently emerged as a promising alternative, offering compact pipelines that map raw visual observations directly to control signals or trajectories. Beyond mitigating cascading errors, such approaches also demonstrate high efficiency and ease of deployment in open-world scenarios. 
However, most approaches mainly replace perception, mapping, and planning modules but still rely on explicit localization modules such as SLAM or visual odometry~\cite{Yang-RSS-23,cai2025navdp} for self-state updates, which requires precise extrinsic calibration between the camera and the robot chassis. This reliance arises because these planners typically process only single frame~\cite{roth2024viplanner} or short clips~\cite{sridhar2024nomad,cai2025navdp}, lacking the ability to summarize long-term histories for consistent state updates. Without temporal grounding, short-term estimates accumulate errors over time, leaving trajectory planning vulnerable to drift and inconsistency.
Also, single frame perception lacks geometric memory needed for robust metric reasoning~\cite{shah2023vint}. Most methods reconstruct only partial or scale-ambiguous geometry, and fails to capture broader spatial context including occluded and rear-view regions, limiting the fidelity of spatial reasoning. 
% Together, these shortcomings prevent current end-to-end systems from achieving robust, metric-aware planning in diverse real-world environments.

\begin{figure}[t]
\begin{center}
  \includegraphics[width=\linewidth]{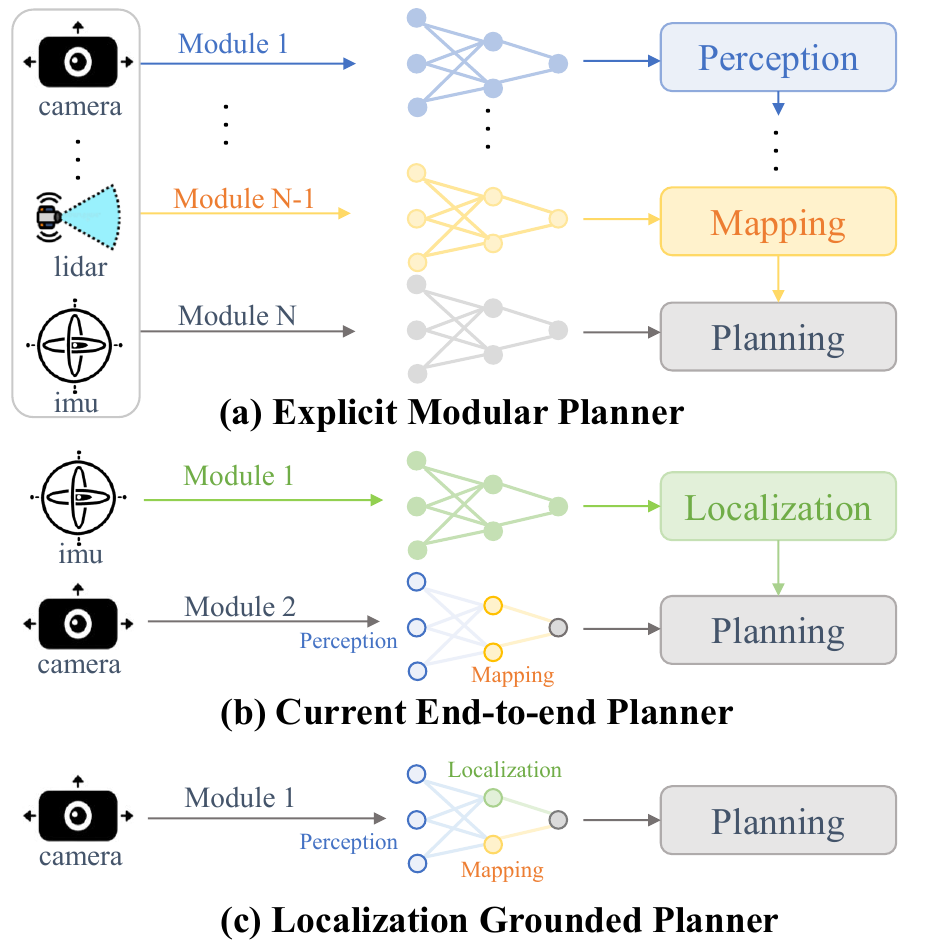}
  \vspace{-0.8cm}
\end{center}
  \caption{(a) Traditional modular planners decompose tasks into modules, introducing cascading errors. (b) Existing end-to-end frameworks directly map observations to control signals but still rely on explicit localization modules. (c) LoGoPlanner integrates implicit state estimation and metric-aware geometry perception into policy for fully end-to-end planning.}
\label{fig:streaming-rate}
\vspace{-0.4cm}
\end{figure}

To address these challenges, we propose LoGoPlanner, a \textbf{lo}calization-\textbf{g}r\textbf{o}unded end-to-end navigation framework that integrates temporal visual geometry estimation with diffusion-based trajectory generation. 

Specifically, to overcome the reliance on explicit localization, we design an implicit state estimation module that operates on long visual sequences. Consecutive image histories are processed using pretrained visual-geometry backbones~\cite{wang2025vggt} for camera extrinsic estimation. We further decouple chassis and camera extrinsics: perception is tied to the camera viewpoint while control is executed at the chassis level. This separation prevents viewpoint-dependent perception errors from propagating into the control space, enabling stable action generation even under varying sensor placements. Training with camera data captured under varying heights and pitch angles enables cross-embodiment and cross-view ego-motion estimation. Additionally, implicit state features are used to project the goal back into the current frame, ensuring consistent goal alignment.
% These features, which capture both historical state and goal information, condition trajectory generation and improve temporal consistency and goal-reaching performance.

Furthermore, we aggregate multi-frame visual features and pose embeddings to generate an implicit reconstruction of the surrounding scene, thereby providing fine-grained geometric priors for planning. Visual geometry backbones such as VGGT~\cite{wang2025vggt} output relative-scale reconstructions, which cannot be directly aligned with planning trajectories. To resolve this, we introduce an efficient post-training procedure that incorporates depth information as a scene-scale prior, enabling the prediction of dense, metric-scale point clouds relative to the robot’s current position. 
% Based on the goal and the perceived environment geometry, the agent generates obstacle-avoiding trajectories toward the goal in the chassis coordinate frame. 
% Finally, we adopt a two-stage training procedure. In the first stage, visual geometry backbones are fine-tuned to be metric-aware, predicting decoupled chassis and camera extrinsics alongside dense scene point clouds. This yields a model capable of metric-aware self-state estimation and environment perception. In the second stage, we perform \textbf{planning-oriented} joint training: features from these auxiliary tasks are used as condition to a diffusion head, which generates trajectories directly in a planning-centered manner. This mitigates error accumulation that would arise from directly using predicted poses or point clouds as inputs.
Inspired by planning-oriented frameworks~\cite{hu2023planning}, we adopt a query-based policy architecture that fuses implicit state\cite{peng2025towards}, geometry, and goal features into a unified planning context. Unlike approaches that pass explicit poses or point clouds downstream and thereby risk error accumulation, our query-driven design enables the diffusion head to operate directly on geometrically grounded representations. This allows the policy to iteratively refine trajectories toward feasible, collision-free solutions, with planning as the ultimate optimization target.

In simulation, LoGoPlanner achieves a 27.3\% relative improvement over baselines that rely on oracle localization, demonstrating the effectiveness of implicit self-state estimation. Furthermore, in diverse real-world scenarios, our framework exhibits robust generalization across different embodiments and environments.

% Its main contributions are summarized as follows:
% Implicit state estimation from long-term visual sequences, leveraging pre-trained visual geometry backbones (e.g., VGGT) to update the robot’s chassis pose over time, jointly encode the initial position and target, and improve temporal consistency and goal-reaching performance in trajectory planning.

% Fine-grained, metric-scale scene reconstruction for robust environment perception, achieved through multi-frame image processing, depth-conditioned post-training, and decoupled camera-chassis extrinsics, providing consistent geometric priors for collision-aware planning and enabling cross-embodiment, cross-view generalization.

% Demonstrated effectiveness and generalization, with over 10\% improvement over baselines using ground-truth localization in simulation, and robust performance in diverse real-world scenarios across different embodiments, viewpoints, and environments.

%% file: article/2.related.tex
\section{Related Work}

\subsection{Learning-based planner}
Recent end-to-end visual navigation frameworks aim to directly map visual inputs to control commands, eliminating the need for traditional modular pipelines. By leveraging semantic and geometric cues, these methods demonstrate the ability to plan across varied terrains. Some prior works adopt a supervised learning paradigm, where robots are trained to imitate expert trajectories or human demonstrations, reasoning about expert actions from vision-based observations~\cite{loquercio2021learning,sadat2020perceive,shah2023vint,wei2025ground}. 
Beyond imitation learning, reinforcement learning has been applied to end-to-end navigation by optimizing policies through trial-and-error in simulation~\cite{wijmans2019dd,xu2025navrl}, but its high sample complexity and sparse rewards make training costly. To overcome these challenges, Yang et al.~\cite{Yang-RSS-23} proposed iPlanner, which reformulates planning as an offline bi-level optimization problem, improving efficiency but only relying on a single frame to capture the robot’s surrounding geometry. 
These methods are still trained in an open-loop manner, predicting the entire trajectory from start to goal. As a result, they lack explicit estimation of intermediate states and typically assume access to metric geometry, treating localization as an external input. Our work builds on these advances by further integrating state estimation and geometry understanding into navigation planning.

\subsection{Video-geometric model}
Recent progress in video-based geometry models~\cite{wang2025vggt,wang2025pi,wang2025continuous,yang2025fast3r} has significantly advanced multi-frame 3D scene understanding. For instance, Video Depth Anything~\cite{chen2025video} extends monocular depth estimation to long video sequences, exploiting temporal information to preserve geometric quality and consistency over time while retaining generalization.
Beyond depth estimation, models such as VGGT~\cite{wang2025vggt} perform full video-based reconstruction. VGGT is a feed-forward neural network that jointly predicts dense 3D attributes, including depth maps, 3D point tracks, and camera extrinsics, from one or more views of a scene. By leveraging long temporal windows, these models provide geometrically consistent reconstructions and explicit camera pose estimates, offering fine-grained priors that are particularly valuable for downstream tasks such as environment perception and navigation~\cite{simeoni2025dinov3}.

\subsection{Monocular Visual Odometry}
% Monocular visual odometry (VO) and SLAM inherently suffer from scale ambiguity. Geometric methods such as MonoSLAM and PTAM rely on handcrafted features and epipolar constraints, but require motion or scene priors and accumulate drift. Direct methods like DSO and LSD-SLAM optimize photometric consistency, reducing dependence on features but still needing scale initialization and being sensitive to illumination changes. Learning-based methods introduce data-driven priors. Depth-assisted approaches like MonoDepth and ORB-SLAM2 recover scale from pseudo-depth, but generalize poorly across domains. End-to-end models such as PoseNet regress pose and scale directly, while MonoRec uses transformers to alleviate drift; however, both remain vulnerable to dynamic or fast-motion scenarios. Fusion-based methods combine multiple cues: BEV-ODOM exploits spatial regularities, CodedVO improves scale via coded apertures, and VINS-Mono integrates IMU data to achieve strong robustness, though at the cost of extra sensors. Most existing approaches depend on external priors—whether geometric assumptions or additional hardware—limiting their scalability and generalization.

Monocular visual odometry (VO) and SLAM methods inherently suffer from scale ambiguity. Classical geometric methods such as MonoSLAM~\cite{davison2007monoslam} and PTAM~\cite{klein2009parallel} rely on handcrafted features and epipolar constraints but require strong motion or scene priors, and are prone to drift accumulation. Direct methods such as DSO~\cite{engel2017direct} and LSD-SLAM~\cite{engel2014lsd} optimize photometric consistency, reducing reliance on features but still needing scale initialization and remaining sensitive to illumination changes. Learning-based methods introduce data-driven priors. Depth-assisted approaches such as MonoDepth~\cite{godard2019digging} and ORB-SLAM2~\cite{mur2015orb} recover scale from monocularly estimated depth, but often generalize poorly across domains. End-to-end networks like PoseNet~\cite{kendall2015posenet} regress pose and scale directly, while transformer-based models such as MonoRec~\cite{wimbauer2021monorec} alleviate drift through long-range dependencies; however, both remain vulnerable to dynamic or fast-motion scenarios. Fusion-based approaches combine multiple cues: BEV-ODOM\cite{wei2024bev} exploits spatial regularities, CodedVO~\cite{shah2024codedvo} improves scale estimation via coded apertures, and VINS-Mono~\cite{qin2018vins} integrates IMU data to enhance robustness, albeit at the cost of additional sensors. Despite these advances, most monocular VO methods still depend on external priors—whether geometric assumptions, scale initialization, or auxiliary hardware—limiting their scalability and generalization.

%% file: article/3.setup.tex
\section{Problem Formulation}

We study the problem of point-goal navigation using only RGB-D observations. An agent must navigate from its start pose to a designated target point \(g \in \mathbb{R}^3\) while avoiding collisions without relying on additional modules.
Time proceeds in discrete steps \(i = 1, \dots, N\). At each step, the agent receives an RGB-D observation \(\mathbf{O}_i = (\mathbf{I}_i, \mathbf{D}_i)\), where \(\mathbf{I}_i \in \mathbb{R}^{H\times W \times 3}\) is the RGB image and \(\mathbf{D}_i \in \mathbb{R}^{H\times W}\) is the corresponding depth map.  
To successfully reach the goal, the agent must continuously estimate its own state from long-term visual history while simultaneously perceiving the surrounding environment to ensure safe navigation.

% For \textbf{geometric perception}, the estimated self-state is used to spatially transform and fuse geometric observations across historical frames, yielding a coherent local representation of the environment. Specifically, the per-frame geometric observation \(P_\tau\) (e.g., point cloud) extracted from \(O_\tau\) is transformed into the current frame as \(\tilde{P}_{\tau \to i} = \mathcal{T}_{\hat{p}_\tau \to \hat{p}_i}\big(P_\tau\big)\), where \(\mathcal{T}_{\hat{p}_\tau \to \hat{p}_i}(\cdot)\) denotes the transformation parameterized by the estimated poses \(\hat{p}_\tau\) and \(\hat{p}_i\).  

% For \textbf{trajectory planning}, the predicted state of the agent up to time \(i\) is given by \(\hat{s}_{1:i} = f(O_{1:i})\), which further enables the transformation of the global goal \(g\) into the current coordinate frame, yielding the relative goal \(\hat{g}_i = f(\hat{s}_{1:i}, g)\). Based on this goal and the perceived environment, the agent induces an obstacle-avoiding trajectory \(\tau_{1:T} = \{p_1, \dots, p_T\}\) toward the goal expressed in the chassis coordinate frame.  
The predicted state of the agent up to time \(i\) is given by \(\hat{s}_{1:i} = f(O_{1:i})\), which further enables the transformation of the global goal \(g\) into the current coordinate frame, yielding the relative goal \(\hat{g}_i = f(\hat{s}_{1:i}, g)\). Based on this goal and the perceived environment, the agent plans an obstacle-avoiding trajectory \(\tau_{1:T} = \{p_1, \dots, p_T\}\) toward the goal expressed in the chassis coordinate frame.  

Unlike existing end-to-end approaches that depend on explicit localization modules, our framework performs implicit, closed-loop state estimation directly from visual sequences. At each step, the agent maintains (i) its estimated chassis pose \(\widehat{p}_i\) in metric scale, (ii) the relative goal position \(\widehat{g}_i\) in the current frame, and (iii) a dense local point cloud \(\widehat{P}_i\) that captures its surrounding environment.

%% file: article/4.method.tex
\section{methodology}

\begin{figure*}[t]
  \centering
  \includegraphics[width=\textwidth]{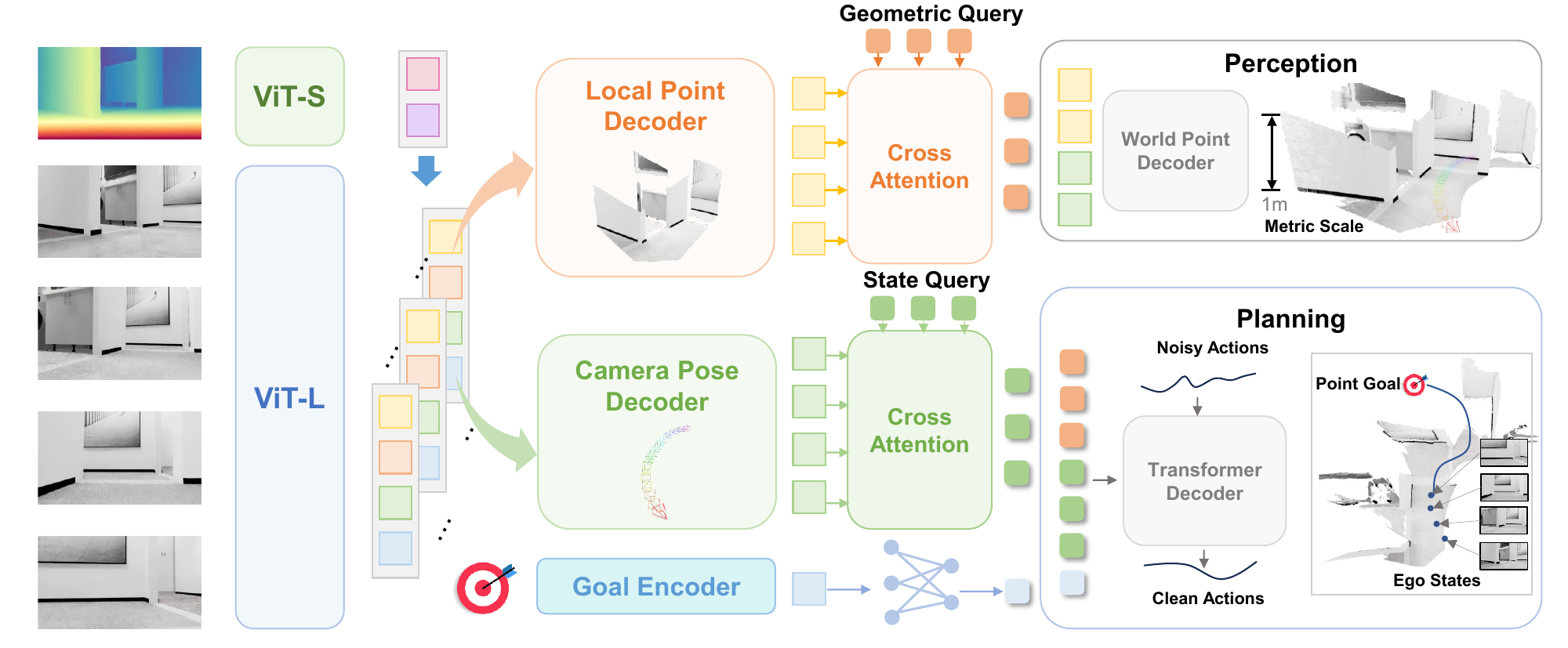}
  \vspace{-0.7cm}
  % \caption{Our method builds on end to-end sparse query-based 3D object detectors, utilizing Latency-aware alignment to integrate historical information and a predictive detection head to enable future state prediction.}
  \caption{\textbf{Architecture overview. }
  Our method injects scale priors into the image patches that are encoded by ViT~\cite{oquab2023dinov2}, and finetunes the video geometry model to metric scale prediction. We adopt a query-based design in which ego state representation and environment geometry are implicitly aggregated through task-specific queries. A diffusion policy head is detached to generate feasible and collision-free trajectories.
  }
  \label{fig:methodology}
  \vspace{-0.5cm}
\end{figure*}

\subsection{Overall Architecture}
As illustrated in Fig.~\ref{fig:methodology}, LoGoPlanner is a unified end-to-end navigation framework that jointly learns metric-aware perception, implicit localization, and trajectory generation. 

The system builds on a pretrained video geometry backbone augmented with depth-derived scale priors.
Through auxiliary supervision on point prediction and pose estimation, the model encodes both fine-grained structures and long-term ego-motion into compact world point embeddings aligned with the planning coordinate system. 
A query-based design allows task-specific queries to extract state and geometry features through cross-attention and fuse them with goal embeddings into a unified planning context. A diffusion-based policy head then conditions on this context to iteratively refine noisy actions into collision-free trajectories. 
% By tightly integrating perception, localization, and planning within a single architecture, LoGoPlanner achieves scalable and calibration-free navigation without reliance on explicit pose estimation.

\subsection{Metric-aware Visual Geometry Learning}
Recent video geometry models such as VGGT~\cite{wang2025vggt} can recover dense 3D scene geometry from image sequences, but their predictions are only defined up to an unknown scale, limiting their applicability to navigation. To address this limitation, we fine-tune the backbone by injecting scale priors from depth maps, enabling metric-scale scene reconstruction.

Concretely, given a causal sequence of $N$ RGB images $\mathbf{I}_{1:N}$ from the same navigation scene, the video geometric model uses a vision transformer~\cite{oquab2023dinov2} to patchify each image $\mathbf{I}_{i}$ into a set of $K$ initial patch tokens $\mathbf{t}^I_i \in \mathbb{R}^{K \times C^I}$. The alternating-attention mechanism alternates between intra-frame and inter-frame attention, improving both local fidelity and long-horizon consistency. 
To inject scale priors into the semantic patches, we employ a lightweight variant of the same vision transformer~\cite{oquab2023dinov2} to encode input depth maps into geometric tokens $\mathbf{t}^D_i \in \mathbb{R}^{K \times C^D}$, which are then fused at the patch level with initial semantic tokens $\mathbf{t}^I_i$. A transformer decoder module, integrating attention mechanisms with Rotary Position Embedding(RoPE)~\cite{su2024roformer}, is further used to integrate information within frames, producing per-frame features that are enriched with metric-scale information. 

\begin{equation}
    \mathbf{t}_i^{metric} = \mathrm{Attention}\big( \mathrm{RoPE}((\mathbf{t}^I_i, \mathbf{t}^D_i), \mathrm{pos}) \big),
\end{equation}
where $\mathrm{pos}\in \mathbb{R}^{K \times 2}$ denotes the spatial position coordinates of image patches, $\mathrm{RoPE}(\cdot, \mathrm{pos})$ applies position-dependent rotations to tokens using 2D coordinates in $\mathrm{pos}$, thereby augmenting the intra-frame attention mechanism to better capture spatial positional relationships between patch tokens, and $\mathbf{t}_i^{metric}$ representing the fused feature embedding at time $i$ with metric-scale awareness.

To improve the accuracy of point cloud prediction, we introduce auxiliary tasks that provide additional supervision during training. Specifically, given the multi-frame feature representation $\mathcal{F}=\{\mathbf{t}_1^{metric},\dots,\mathbf{t}_N^{metric}\}$ from the context RGB frames, we branch the features into two task-specific heads: a local point head and a camera pose head. 

\textbf{Local point prediction.}  
For each frame $i$, the local point head $\phi_p$ maps metric-aware tokens $\mathbf{t}_i^{metric}$ to a latent feature representation $\mathbf{h}^p_i$, which is further decoded to predict canonical local 3D points in the camera coordinate system:
\begin{equation}
\mathbf{h}^p_i = \phi_p(\mathbf{t}_i^{metric}), 
\qquad 
\widehat{P}^{\mathrm{local}}_i = f_p(\mathbf{h}^p_i)
\end{equation}
where $\widehat{P}^{\mathrm{local}}_i = \{\hat{\mathbf{p}}^{\mathrm{local}}_{i,j}\}_{j=1}^{M}$ denotes a set of predicted local points $j$ for frame $i$, which is supervised by local points in the camera coordinate system using the pinhole model for each pixel $(u,v)$ in images:
\begin{equation}
\mathbf{p}_{\mathrm{cam},i}(u,v) \;=\; D_i(u,v)\,K^{-1}[u\; v\; 1]^\top
\end{equation}
Unlike directly using noisy depth maps for local point projection, our data-driven prediction improves reconstruction robustness and provides implicit features for world point prediction.

\textbf{Camera pose prediction.}  
In parallel, the camera pose head $\phi_\mathrm{c}$ maps $\mathbf{t}_i^{metric}$ to another task-specific feature $\mathbf{h}^c_i$, which is decoded into a predicted camera-to-world transformation:
\begin{equation}
\mathbf{h}^\mathrm{c}_i = \phi_\mathrm{c}(\mathbf{t}_i^{metric}), 
\qquad 
\widehat{T}_{\mathrm{c},i} = f_c(\mathbf{h}^\mathrm{c}_i)
\end{equation}
where the world coordinate system is defined with respect to the chassis frame of the last time step, ensuring consistency between the predicted camera trajectory and the planning coordinate system.

\textbf{World point prediction.}  
Rather than directly encoding the predicted local points or poses, we exploit their intermediate task-specific features to perform implicit transformations, thereby obtaining a compact representation of world points for downstream planning. For each frame $i$, we concatenate the local-point feature $\mathbf{h}^p_i$ and the pose feature $\mathbf{h}^\mathrm{c}_i$.
These per-frame fused features are then aggregated across the patches by a context fusion module $\mathcal{A}$:
\begin{equation}
\mathbf{h}_i^w = \mathcal{A}([\mathbf{h}^\mathrm{p}_i,\mathbf{h}^\mathrm{c}_i]).
\end{equation}
Finally, the aggregated representation $\mathbf{h}_i^w$ is passed through a point-cloud decoder $\psi$ to upsample to the target resolution. The intermediate output $\mathbf{z}_i$ is further processed to enforce a scale-invariant range and preserve sign information:
\begin{equation}
\mathbf{z}_i = \psi(\mathbf{h}_i^w),
\qquad 
\widehat{P}^{\mathrm{world}}_i = \mathrm{sign}(\mathbf{z}_i) \cdot (\mathrm{exp}(|\mathbf{z}_i|) -1)
\end{equation}
where $\mathrm{exp}(|\cdot|)-1$ is used to enhance expressiveness of large coordinate values while avoiding saturation, $\widehat{P}^{\mathrm{world}}_i$ represents the reconstructed metric-scale scene point cloud expressed in the last frame to align the coordinate with planning trajectory.

\subsection{Localization Grounded Navigation Policy}
% In the perception process, the observed viewpoint is obtained through the projection of the camera pose, while in the planning process, the controlled object during robot motion is the chassis. This leads to a discrepancy between the two. In cross-embodiment and cross-view scenarios, where the camera height and pitch angle may vary arbitrarily, traditional methods typically require extrinsic calibration between the camera and the chassis to achieve precise control by aligning the perception results with the planning module.  
In navigation, perception is tied to the camera viewpoint, whereas control is executed at the chassis level. This mismatch leads to alignment errors, particularly in cross-embodiment settings where camera height and pitch may vary significantly. Traditional methods rely on explicit extrinsic calibration between the camera and chassis to align perception with planning, but such calibration is fragile and fails to generalize across different views.

In contrast, we decouple the estimation of camera and chassis poses into separate prediction tasks and leverage implicit feature interaction to bridge perception and planning without explicit calibration. Specifically, we assume the robot-mounted camera has no yaw rotation relative to the chassis. The model also predicts the chassis pose and latest goal position from pose estimation task-specific feature $\mathbf{h}^c_i$:
\begin{equation}
\widehat{T}_{\mathrm{b},i} = f_b(\mathbf{h}^\mathrm{c}_i)
\qquad
\widehat{g}_{i} = f_g(\mathbf{h}^\mathrm{c}_i, g)
\end{equation}
which is defined on the ground plane as $(x_i, y_i, \theta_i)$, where $(x_i, y_i)$ denotes the current position relative to the starting point, and $\theta_i$ is the yaw angle.  
% For the camera pose, in order to better align with the current decision trajectory, we use the latest chassis frame as the reference and predict the camera trajectory from the start point up to the current position. 
Formally, the camera pose at time $i$ is obtained from the chassis pose through the extrinsic transformation:
\begin{equation}
T_{\mathrm{b},i} = T_{\mathrm{c},i} \cdot T_{\text{ext}},
\end{equation}
where $T_{\mathrm{b},i}$ is the chassis transformation with respect to the start point, and $T_{\text{ext}}$ denotes the fixed extrinsic transformation capturing the camera’s height and pitch angle relative to the chassis.  
To endow the model with robustness across embodiments, we construct training data under arbitrary camera heights and varying pitch angles, thereby enabling the system to generalize across diverse camera configurations.

% End-to-end planning via implicit feature interaction:
To achieve end-to-end planning, our approach does not explicitly feed the predicted extrinsics or point clouds into the network for goal transformation or trajectory optimization. Instead, inspired by UniAD~\cite{hu2023planning}, we adopt a \textbf{query-based design} in which different modules are aggregated through task-specific queries. Interactions across modules are realized via query cross attention.
We set state queries $Q_s$ to extract implicit state representation from pose specific tokens and geometric queries $Q_g$ to extract implicit environment geometry from world point specific tokens, thus providing sufficient information for trajectory planning:
\begin{equation}
Q_S = \mathrm{CrossAttn}(Q_s, \mathbf{h}^\mathrm{c})
\end{equation}
\begin{equation}
Q_G = \mathrm{CrossAttn}(Q_g, \mathbf{h}^\mathrm{p})
\end{equation}
where the generated $Q_S$, $Q_G$ and goal embedding are concatenated and passed to transformer decoder to produce the planning context query $Q_{P}$:
% \begin{equation}
% Q_G = \mathrm{SelfAttn}([Q_S, Q_g])
% \end{equation}
These implicit features, which encode states and geometric properties, serve as conditioning signals for planning. This strategy avoids cascading errors that would otherwise arise from directly applying upstream predictions to downstream tasks, while ensuring that the final optimization target remains the trajectory planning error.  

% Two-stage training:
For navigation trajectory planning, we attach a diffusion policy head to generate action chunks $\{\boldsymbol{a}_t=(\Delta x_t,\Delta y_t,\Delta\theta_t)\}_{t=1}^T$. Starting from $\boldsymbol{a}^K$ sampled from Gaussian noise, the model predicts noise from noisy action sequences, performs $K$ iterative steps of denoising to produce a series of intermediate actions with decreasing levels of noise:
\begin{equation}
\boldsymbol{a}^{k-1} = \alpha(\boldsymbol{a}^k-\gamma\epsilon_\theta(Q_P, \boldsymbol{a}^{k},k)+\mathcal{N}(0,\sigma^2I))
\end{equation}
where \(\boldsymbol{a}^k\) denotes the noisy action at step \(k\), \(\epsilon_\theta\) is the noise prediction network conditioned on planning context query $Q_{P}$, $\alpha$ and $\gamma$ are the standard diffusion schedule parameters. This formulation enables the policy to iteratively refine actions toward feasible, collision-free trajectories.

%% file: article/5.experiment.tex
\section{experiments}

\subsection{Datasets and Implementation Details}
We use a large-scale navigation dataset~\cite{interndata_n1}, generated with a simulation pipeline designed to efficiently generate diverse robot trajectories across a variety of 3D environments. The robot is modeled as a cylindrical rigid body with a differential-drive two-wheel configuration and is equipped with an RGB-D camera mounted on the top. To simulate embodiment variations across different robotic platforms, the robot’s height is randomized between 0.25 m and 1.25 m, while the camera’s pitch angle is randomized between 0° and 30°. Initial paths between randomly sampled start and goal points are generated using the A* algorithm. These paths are refined through greedy search and subsequently smoothed via cubic spline interpolation to ensure collision-free navigation. The dataset comprises over 200k trajectories and approximately 10 million rendered images.

We adopt a two-stage training paradigm. In the first stage, we fine-tune the decoder of the video geometry model and the task-specific head with a batch size of 12 for 24 hours. During this process, depth-based scale priors are injected, and supervision is provided by metric-scale scene point clouds and camera extrinsics. In the second stage, we jointly train the diffusion head together with the task-specific head while keeping the decoder of the backbone frozen. This stage uses a batch size of 32 and runs for three days, ensuring both robust state estimation and stable perception capabilities.

% We assess our method through both simulated and real world evaluations, comparing it to various learning-based baselines.  

% \subsection{Implementation Details}
\subsection{Main Results}
We evaluate the performance of different learning-based planners in both simulation and real-world environments. The test environments are unseen during training, requiring the robot to continuously estimate its state, plan collision-free trajectories, and navigate toward the designated goal. Planning performance is measured using Success Rate (SR) and Success weighted by Path Length (SPL). The experimental results are summarized in Table~\ref{tab:simulation-results} and Table~\ref{tab:realworld-results}.
\subsubsection{Simulation Experiments}

To simulate realistic environments, we selected 40 scenes from the InternScenes dataset~\cite{grutopia}, including 20 home and 20 commercial scenes. Home scenes are characterized by narrow passages and cluttered semantic layouts, while commercial scenes cover representative categories such as hospitals, supermarkets, restaurants, schools, libraries, and offices. In each scene, 100 start–goal pairs are randomly sampled in unoccupied spaces with distances of 4–10 meters, and initial orientations are determined through path planning to avoid collisions.

\begin{figure}[t]
\begin{center}
  \includegraphics[width=\linewidth]{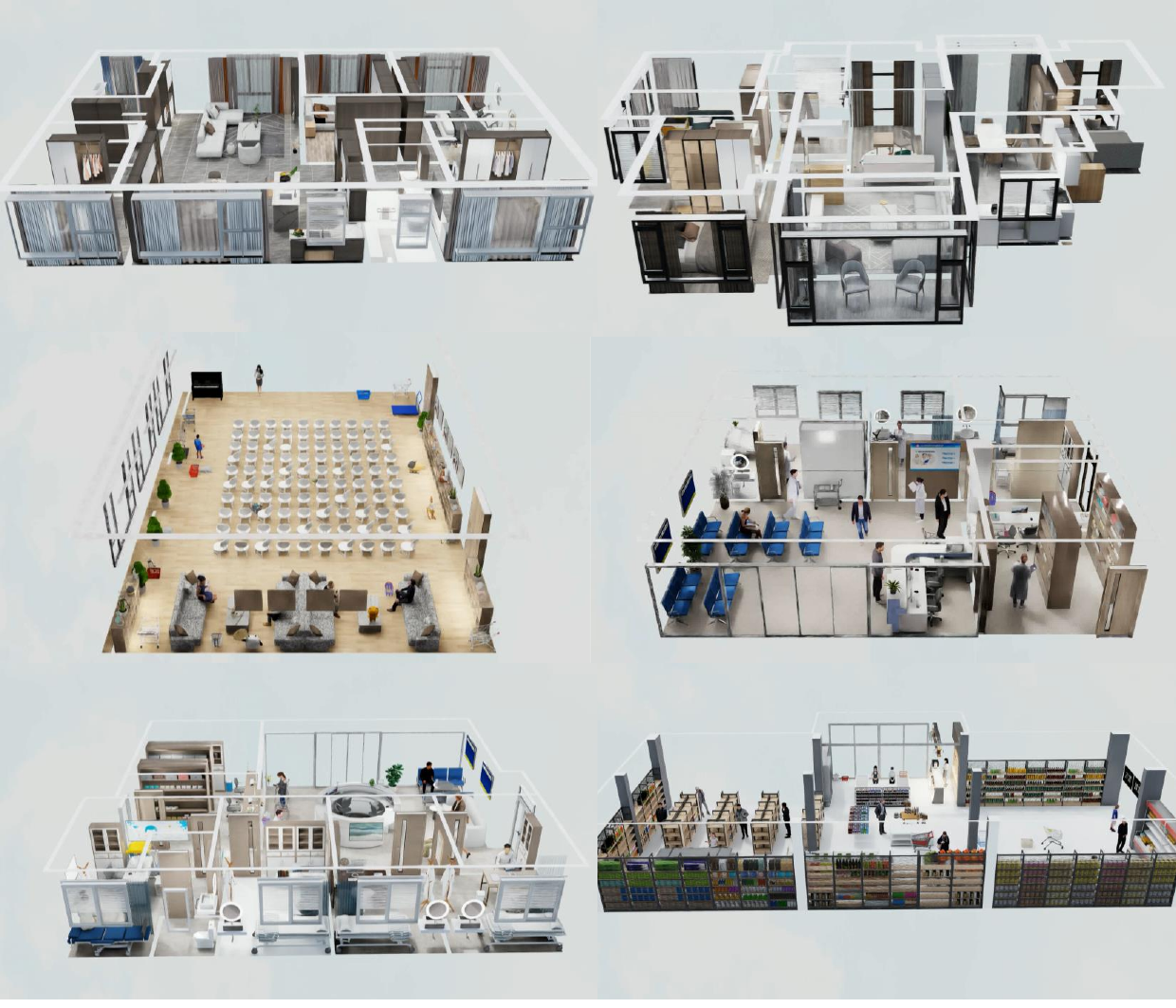}
  \vspace{-0.8cm}
\end{center}
  \caption{Home scenes are characterized by narrow passages and cluttered semantic layouts, while commercial scenes cover representative categories such as hospitals, supermarkets, restaurants, schools, libraries, and offices.}
\label{fig:scene}
\vspace{-0.1cm}
\end{figure}

% ---------------- Simulation ----------------
\begin{table}[ht]
  \centering
  \caption{Simulation Results}
  \begin{threeparttable}
  \begin{tabular}{c|c|cc|cc}
      \toprule
      \multirow{2}{*}{\textbf{Planner}} & \multirow{2}{*}{\textbf{Localization}} & \multicolumn{2}{c|}{\textbf{Home}} & \multicolumn{2}{c}{\textbf{Commercial}} \\
      \cmidrule(lr){3-6}
      & & \textbf{SR$\uparrow$} & \textbf{SPL$\uparrow$} & \textbf{SR$\uparrow$} & \textbf{SPL$\uparrow$} \\
      \midrule
      \multirow{2}{*}{DD-PPO~\cite{wijmans2019dd}} & $\times$ & -- & -- & -- & -- \\
                              & $\checkmark$ & 0.4 & 0.4 & 5.3 & 5.2 \\
      \midrule
      \multirow{2}{*}{iPlanner~\cite{Yang-RSS-23}} & $\times$ & 41.7 & 40.2 & 53.1 & 51.8 \\
                                & $\checkmark$ & 43.0 & 40.6 & 54.6 & 52.8 \\
      \midrule
      \multirow{2}{*}{ViPlanner~\cite{roth2024viplanner}} & $\times$ & 44.0 & 42.8 & 61.3 & 60.1 \\
                                 & $\checkmark$ & 45.0 & 43.2 & 63.7 & 61.9 \\
      \midrule
      LoGoPlanner & $\times$ & \textbf{57.3} & \textbf{52.4} & \textbf{67.1} & \textbf{63.9} \\
      \bottomrule
  \end{tabular}
  \begin{tablenotes}
    \footnotesize
    \item[$\times$] With explicit or implicit localization.
    \item[$\checkmark$] With oracle localization from simulator.
  \end{tablenotes}
  \end{threeparttable}
  \label{tab:simulation-results}
  \vspace{-0.7cm}
\end{table}

\begin{figure*}[t]
\begin{center}
  \includegraphics[width=\linewidth]{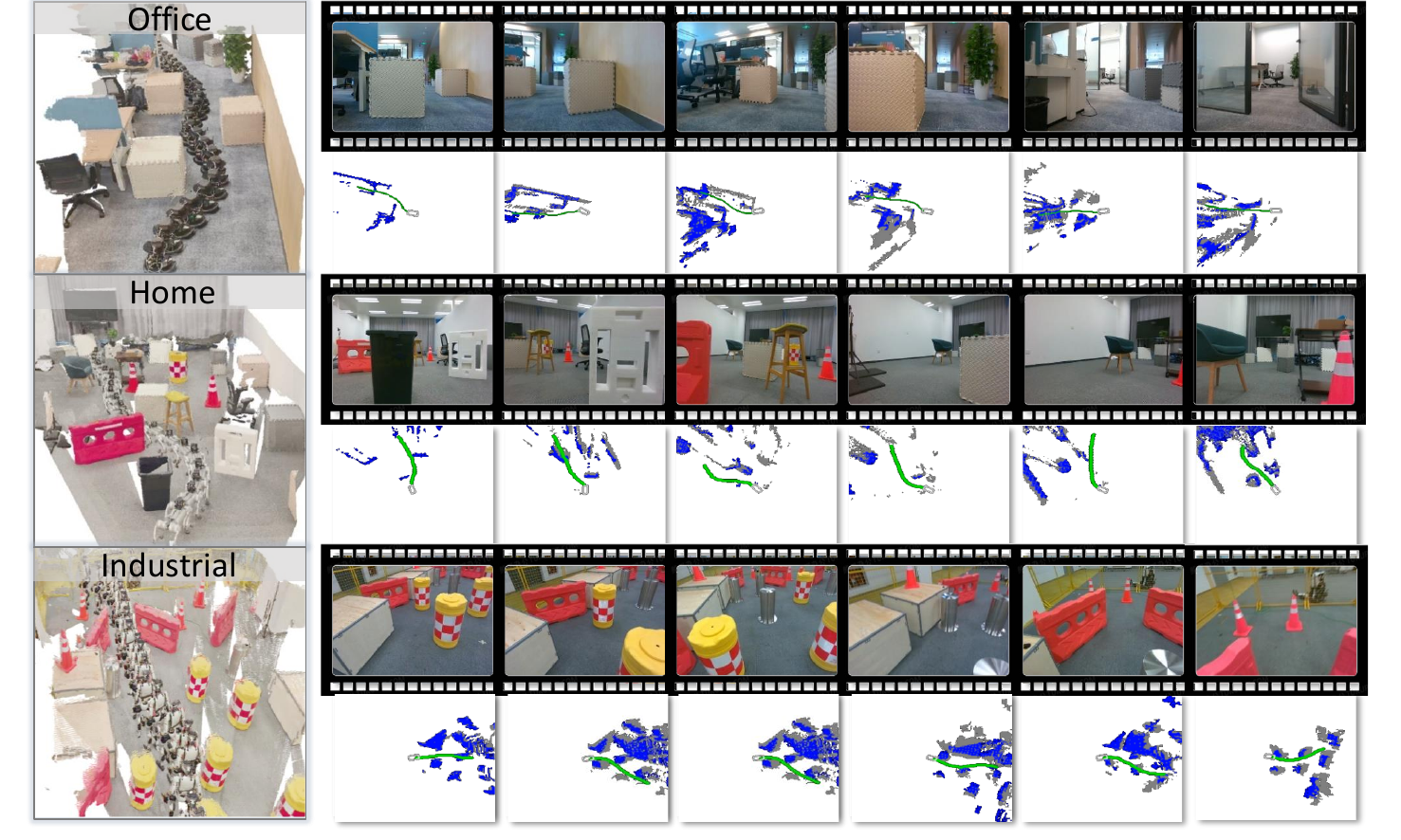}
  \vspace{-1.0cm}
\end{center}
  \caption{Visualization of LoGoPlanner in real-world scenarios on different robot platforms. The green curves are the planned trajectories of LoGoPlanner. Blue and grey clouds are the obstacles of the current frame and the previous frame respectively.}
\label{fig:streaming-rate}
\vspace{-0.4cm}
\end{figure*}

Table~\ref{tab:simulation-results} reports navigation performance for two scene categories. In the table, the “Localization” column indicates whether the planner has access to ground-truth localization from the simulator. A “$\checkmark$” means the planner uses simulator-provided poses, while a “$\times$” indicates no access to ground-truth localization: for iPlanner and ViPlanner, an external visual odometry module (ORB-SLAM3~\cite{ORBSLAM3_TRO} with RGB-D input) is used, whereas LoGoPlanner performs implicit state estimation without any external localization.

% LoGoPlanner, which integrates metric-aware self-localization with geometry-aware planning from visual inputs, consistently achieves the best performance. 
Reinforcement-learning-based planners like DD-PPO~\cite{wijmans2019dd} typically require massive interaction data and careful reward shaping. They can underperform in unseen environments due to overfitting to training distributions and sparse, noisy reward signals. 
The rule-based nature of imperative planning~\cite{Yang-RSS-23,roth2024viplanner} leads to failure to adjust to novel spatial configurations. Moreover, both methods operate on single-frame input, which restricts their ability to capture holistic scene geometry and leads to poor adaptation in cluttered or unstructured settings.

In contrast, LoGoPlanner achieves stronger robustness by jointly incorporating ego-state information and multi-frame geometric reconstruction. This design ensures greater consistency in trajectory generation while providing richer spatial perception, which in turn enhances obstacle avoidance and overall navigation performance. Compared with baselines that benefit from external or oracle localization, LoGoPlanner improves Home SR by 27.3\% points and Home SPL by 21.3\% relative to ViPlanner, demonstrating that integrating implicit self-localization with geometry-aware planning yields superior closed-loop navigation.

\subsubsection{Real-world Experiments}
To evaluate the cross-platform, cross-scene, and cross-view generalization of LoGoPlanner in real-world settings, we deploy the system on three distinct robotic platforms under diverse environment configurations. For quantitative evaluation of vision-based navigation methods in real-world scenarios, we test a TurtleBot in an office scene with structured obstacles, a Unitree Go2 in a cluttered home scene containing arbitrarily shaped obstacles, and a Unitree G1 in an industrial scene with road-block obstacles, evaluating 20 trajectories. Quantitative results are shown in Table \ref{tab:realworld-results}. All algorithms run on an NVIDIA RTX 4090 GPU for cloud-based inference, with control commands transmitted to the robots in real time.

iPlanner performs relatively poorly across different embodiments because its trajectory inconsistency during obstacle avoidance often leads to collisions. In contrast, ViPlanner demonstrates better performance; however, constrained by the training data and network design of single-frame-based navigation policy, it exhibits poor performance in challenging scenarios (e.g., industrial environments with Unitree G1).

LoGoPlanner can be deployed directly without requiring visual odometry or SLAM. Despite camera jitter caused by the quadruped platform, LoGoPlanner achieves accurate self-localization and generates reliable collision-free trajectories towards the goal. By leveraging point clouds as implicit intermediate representations, the model further reduces the sim-to-real gap, demonstrating the framework’s robust generalization, reduced deployment complexity, and readiness for direct application in scenarios with varying scene structure and camera viewpoints. For detailed demonstrations of our system, please refer to the \textcolor{blue}{\textit{demo video}}.

% ---------------- Real World ----------------
\begin{table}[ht]
  \centering
  \caption{Real-World Results}
  \begin{tabular}{c|c|c|c}
      \toprule
      \multirow{3}{*}{\textbf{Planner}} & \textbf{Office} & \textbf{Home} & \textbf{Industrial} \\
      \cmidrule(lr){2-4}
      & \small{TurtleBot} & \small{Unitree Go2} & \small{Unitree G1} \\
      \cmidrule(lr){2-4}
      & \textbf{SR$\uparrow$} & \textbf{SR$\uparrow$} & \textbf{SR$\uparrow$} \\
      \midrule
      iPlanner~\cite{Yang-RSS-23} & 10 (2/20) & 15 (3/20) & 0 (0/20) \\
      ViPlanner~\cite{roth2024viplanner} & 50 (10/20) & 45 (9/20) & 0 (0/20) \\
      LoGoPlanner & 85 (17/20) & 70 (14/20) & 50 (10/20) \\
      \bottomrule
  \end{tabular}
  \label{tab:realworld-results}
  \vspace{-0.4cm}
\end{table}

\begin{figure*}[t]
\begin{center}
  \includegraphics[width=\linewidth]{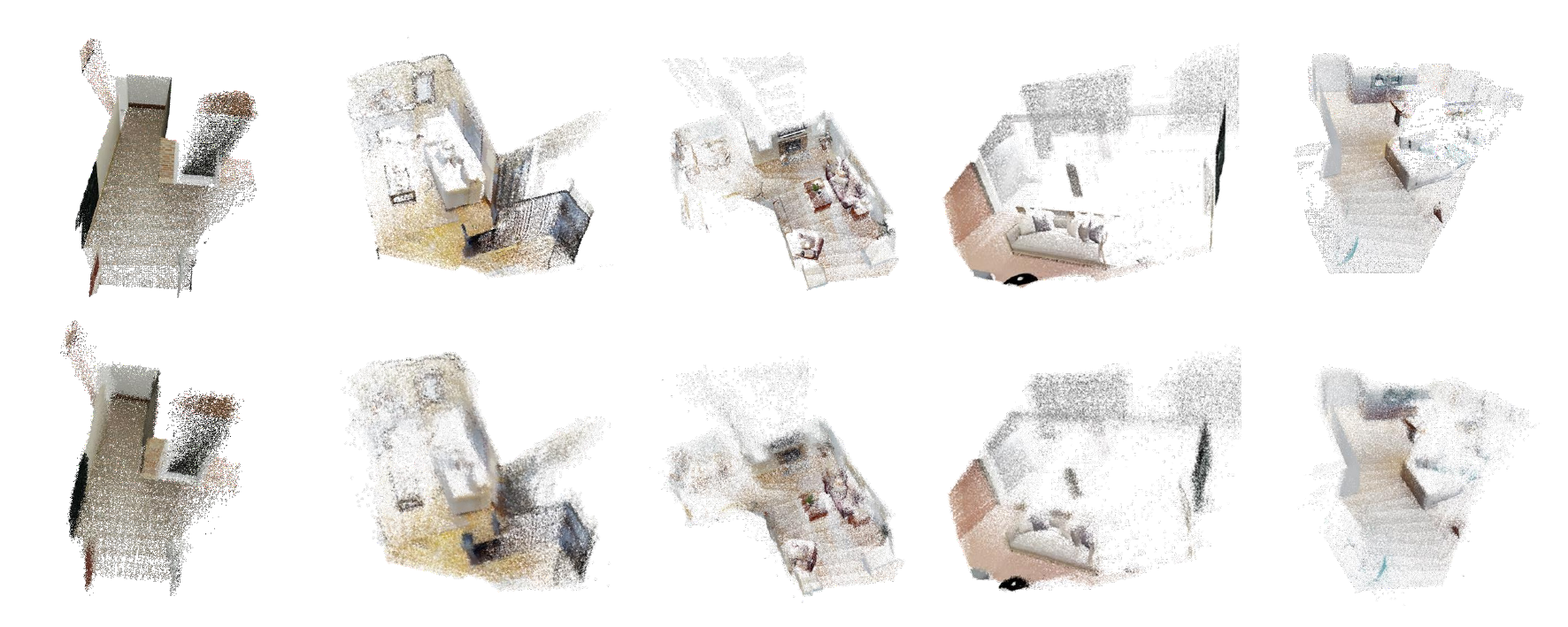}
  \vspace{-1.0cm}
\end{center}
  \caption{Visualization of reconstruction results: the first row shows the scene point cloud of the ground truth, and the second row shows the predicted scene point cloud. The point cloud at the metric scale is predicted with the chassis of the last frame as the coordinate origin.}
\label{fig:streaming-rate}
\vspace{-0.4cm}
\end{figure*}

\subsection{Ablation Study}
To endow the model with self-state estimation and metric-aware perception, we introduce three auxiliary tasks in the first stage. \textbf{Odometry} supervises ego-motion estimation, \textbf{Goal} provides dynamic target updates based on self-state, and \textbf{Point Cloud} transforms historical image observations into geometric point clouds using estimated camera extrinsics. 
% Without these tasks, end-to-end supervision yields basic path-planning ability, but intermediate features lack guidance and targets can be lost under inaccurate state estimation. Adding odometry and goal supervision improves ego-motion perception and trajectory consistency, while further point cloud supervision captures obstacle spatial relations beyond 2D semantics, enhancing obstacle avoidance.
Without any auxiliary tasks, the model trained solely with end-to-end supervision exhibits basic path-planning capability. However, intermediate features lack explicit guidance, and the model may lose track of the goal when self-state estimation is inaccurate. Incorporating \textbf{Odometry} and \textbf{Goal} supervision improves ego-motion estimation and enhances trajectory consistency. Further adding \textbf{Point Cloud} supervision allows the model to capture the spatial relationships of obstacles beyond 2D semantics, significantly improving obstacle avoidance in trajectory generation.

\begin{table}[h]
  \centering
  \caption{Ablation Experiments on Key Modules}
  % 7列结构：3模块列 + 2 Home指标列 + 2 Commercial指标列，单栏适配无溢出
  \begin{tabular}{ccc|cc|cc}
    \toprule
    % 顶层表头：Modules跨3列，Home/Commercial各跨2列，竖线分隔场景
    \multicolumn{3}{c|}{\textbf{Modules}} & \multicolumn{2}{c|}{\textbf{Home}} & \multicolumn{2}{c}{\textbf{Commercial}} \\
    \cmidrule(lr){1-7} % 分割顶层与底层表头，线条不超出表格
    % 底层表头：具体模块名与指标名
    \textbf{Odometry} & \textbf{Goal} & \textbf{Point Cloud} & \textbf{SR$\uparrow$} & \textbf{SPL$\uparrow$} & \textbf{SR$\uparrow$} & \textbf{SPL$\uparrow$} \\
    \midrule
    $\times$      & $\times$         & $\times$             & 49.5 & 47.0 & 59.4 & 57.0 \\ % All Modules Disabled
    $\checkmark$  & $\times$         & $\times$             & 51.3 & 49.7 & 61.2 & 59.2 \\ % Only Odom Module
    $\checkmark$  & $\checkmark$     & $\times$             & 52.4 & 50.1 & 63.3 & 60.3 \\ % Odom + Subgoal Modules
    $\checkmark$  & $\checkmark$     & $\checkmark$         & \textbf{57.3} & \textbf{52.4} & \textbf{67.1} & \textbf{63.9} \\ % All Modules Enabled
    \bottomrule
  \end{tabular}
  \label{tab:ablation_both_scenes}
  \vspace{-0.2cm}
\end{table}

Our model grounds the pose estimation and reconstruction capabilities of the video geometry model into the planning policy, making the choice of geometric backbone crucial for the task. We experimented with several backbones: single-frame geometric backbones, DepthAnything~\cite{yang2024depth}, multi-frame geometric backbones, Video DepthAnything~\cite{chen2025video}, VGGT$\dagger$~\cite{wang2025vggt} without metric scale, and a scale-injected version of VGGT~\cite{wang2025vggt}. During these experiments, all auxiliary tasks are retained in training.

To quantify pose estimation and planning accuracy, we define two metrics: \textbf{Navigation Error} (NE), the Euclidean distance between the robot’s final stopping position and the goal, and \textbf{Planning Error} (PE), the distance between the endpoint of the planned trajectory and the goal.
Single-frame backbones provide per-frame depth perception to for obstacle avoidance but lack temporal consistency. Multi-frame pretrained geometric models capture temporal correlations, yet without supervision on camera poses, they struggle to model accurate sequential pose relationships and keep planning consistency. Existing reconstruction-pretrained models offer reliable ego-motion estimation; however, after fine-tuning without depth prior, the estimated camera poses follow the correct trends but exhibit scale discrepancies. After incorporating scale priors, our model not only achieves higher planning success rates but also demonstrates improved planning accuracy. Therefore, metric-scale supervision is still required for real-world applications.

\begin{table}[h]
\centering
\caption{Performance under Different Video Geometry Backbones}
  \begin{threeparttable}
  \resizebox{\linewidth}{!}{
\begin{tabular}{c|cccc|cccc}
\toprule
\multirow{2}{*}{\textbf{Backbone}} 
& \multicolumn{4}{c|}{\textbf{Home}} 
& \multicolumn{4}{c}{\textbf{Commercial}} \\
\cmidrule(lr){2-5} \cmidrule(lr){6-9} 
~ & \textbf{SR$\uparrow$} & \textbf{SPL$\uparrow$} & \textbf{NE$\downarrow$} & \textbf{PE$\downarrow$} 
& \textbf{SR$\uparrow$} & \textbf{SPL$\uparrow$} & \textbf{NE$\downarrow$} & \textbf{PE$\downarrow$} \\
\midrule
DA~\cite{yang2024depth} & 49.9 & 47.1 & 2.51 & 1.48 & 59.9 & 57.4 & 2.49 & 1.49 \\
VDA~\cite{chen2025video} & 51.5 & 48.2 & 2.43 & 1.08 & 61.4 & 58.8 & 2.15 & 1.08 \\
VGGT$\dagger$~\cite{wang2025vggt} & 54.9 & 50.4 & 2.35 & 0.87 & 62.4 & 57.7 & 2.31 & 1.18 \\
VGGT~\cite{wang2025vggt} & \textbf{57.3} & \textbf{52.4} & \textbf{2.24} & \textbf{0.55} & \textbf{67.1} & \textbf{63.9} & \textbf{2.07} & \textbf{0.59} \\
\bottomrule
\end{tabular}}
  \begin{tablenotes}
    \footnotesize
    \item[$\dagger$]: without injecting depth as the scale prior.
  \end{tablenotes}
  \end{threeparttable}
\label{tab:backbone_performance}
\end{table}

%% file: article/6.conclusion.tex
\section{CONCLUSIONS}

We proposed LoGoPlanner, a localization-grounded end-to-end navigation framework that unifies metric-aware pose estimation, long-horizon scene reconstruction, and feature-level policy conditioning. By integrating implicit self-state estimation with fine-grained environmental perception, our method overcomes limitations of traditional modular pipelines and prior end-to-end approaches that rely on external localization. Experiments in both simulation and real-world scenarios demonstrate that LoGoPlanner achieves superior trajectory planning and obstacle avoidance, while generalizing robustly across diverse embodiments, viewpoints, and environments. This work highlights the potential of grounding navigation policies in geometric and metric-aware priors, pointing toward more autonomous, reliable, and adaptable robotic navigation in unstructured real-world settings.

% A key limitation of LoGoPlanner is that imitation learning on offline static trajectories limits its ability to handle highly dynamic scenes. Integrating online adaptation, such as 3D point tracking~\cite{karaev2024cotracker,peng2025towards} or reinforcement learning-based fine-tuning~\cite{ren2024diffusion}, could improve responsiveness to moving obstacles and evolving environments.

Due to the limited number($\sim$2k) of available navigation scenes, the reconstruction performance in real-world environments remains unsatisfactory. We are currently training on real world datasets in metric-scale, to enhance the performance for practical deployment.

%% file: root.bib
@article{loquercio2021learning,
  title={Learning high-speed flight in the wild},
  author={Loquercio, Antonio and Kaufmann, Elia and Ranftl, Ren{\'e} and M{\"u}ller, Matthias and Koltun, Vladlen and Scaramuzza, Davide},
  journal={Science Robotics},
  volume={6},
  number={59},
  pages={eabg5810},
  year={2021},
  publisher={American Association for the Advancement of Science}
}

@inproceedings{sadat2020perceive,
  title={Perceive, predict, and plan: Safe motion planning through interpretable semantic representations},
  author={Sadat, Abbas and Casas, Sergio and Ren, Mengye and Wu, Xinyu and Dhawan, Pranaab and Urtasun, Raquel},
  booktitle={European Conference on Computer Vision},
  pages={414--430},
  year={2020},
  organization={Springer}
}

@inproceedings{shah2023vint,
  title={ViNT: A Foundation Model for Visual Navigation},
  author={Shah, Dhruv and Sridhar, Ajay and Dashora, Nitish and Stachowicz, Kyle and Black, Kevin and Hirose, Noriaki and Levine, Sergey},
  booktitle={Conference on Robot Learning},
  pages={711--733},
  year={2023},
  organization={PMLR}
}

@INPROCEEDINGS{Yang-RSS-23, 
    AUTHOR    = {Fan Yang AND Chen Wang AND Cesar Cadena AND Marco Hutter}, 
    TITLE     = {{iPlanner: Imperative Path Planning}}, 
    BOOKTITLE = {Proceedings of Robotics: Science and Systems}, 
    YEAR      = {2023}, 
    ADDRESS   = {Daegu, Republic of Korea}, 
    MONTH     = {July}, 
    DOI       = {10.15607/RSS.2023.XIX.064} 
}

@inproceedings{hu2023planning,
  title={Planning-oriented autonomous driving},
  author={Hu, Yihan and Yang, Jiazhi and Chen, Li and Li, Keyu and Sima, Chonghao and Zhu, Xizhou and Chai, Siqi and Du, Senyao and Lin, Tianwei and Wang, Wenhai and others},
  booktitle={Proceedings of the IEEE/CVF conference on computer vision and pattern recognition},
  pages={17853--17862},
  year={2023}
}

@article{yang2024depth,
  title={Depth anything v2},
  author={Yang, Lihe and Kang, Bingyi and Huang, Zilong and Zhao, Zhen and Xu, Xiaogang and Feng, Jiashi and Zhao, Hengshuang},
  journal={Advances in Neural Information Processing Systems},
  volume={37},
  pages={21875--21911},
  year={2024}
}

@inproceedings{chen2025video,
  title={Video depth anything: Consistent depth estimation for super-long videos},
  author={Chen, Sili and Guo, Hengkai and Zhu, Shengnan and Zhang, Feihu and Huang, Zilong and Feng, Jiashi and Kang, Bingyi},
  booktitle={Proceedings of the Computer Vision and Pattern Recognition Conference},
  pages={22831--22840},
  year={2025}
}

@inproceedings{wang2025vggt,
  title={Vggt: Visual geometry grounded transformer},
  author={Wang, Jianyuan and Chen, Minghao and Karaev, Nikita and Vedaldi, Andrea and Rupprecht, Christian and Novotny, David},
  booktitle={Proceedings of the Computer Vision and Pattern Recognition Conference},
  pages={5294--5306},
  year={2025}
}

@article{wijmans2019dd,
  title={Dd-ppo: Learning near-perfect pointgoal navigators from 2.5 billion frames},
  author={Wijmans, Erik and Kadian, Abhishek and Morcos, Ari and Lee, Stefan and Essa, Irfan and Parikh, Devi and Savva, Manolis and Batra, Dhruv},
  journal={arXiv preprint arXiv:1911.00357},
  year={2019}
}

@inproceedings{roth2024viplanner,
  title={Viplanner: Visual semantic imperative learning for local navigation},
  author={Roth, Pascal and Nubert, Julian and Yang, Fan and Mittal, Mayank and Hutter, Marco},
  booktitle={2024 IEEE International Conference on Robotics and Automation (ICRA)},
  pages={5243--5249},
  year={2024},
  organization={IEEE}
}

@inproceedings{cao2022autonomous,
  title={Autonomous exploration development environment and the planning algorithms},
  author={Cao, Chao and Zhu, Hongbiao and Yang, Fan and Xia, Yukun and Choset, Howie and Oh, Jean and Zhang, Ji},
  booktitle={2022 International Conference on Robotics and Automation (ICRA)},
  pages={8921--8928},
  year={2022},
  organization={IEEE}
}

@article{fan2021step,
  title={Step: Stochastic traversability evaluation and planning for risk-aware off-road navigation},
  author={Fan, David D and Otsu, Kyohei and Kubo, Yuki and Dixit, Anushri and Burdick, Joel and Agha-Mohammadi, Ali-Akbar},
  journal={arXiv preprint arXiv:2103.02828},
  year={2021}
}

@inproceedings{wellhausen2021rough,
  title={Rough terrain navigation for legged robots using reachability planning and template learning},
  author={Wellhausen, Lorenz and Hutter, Marco},
  booktitle={2021 IEEE/RSJ International Conference on Intelligent Robots and Systems (IROS)},
  pages={6914--6921},
  year={2021},
  organization={IEEE}
}

@inproceedings{gog2021pylot,
  title={Pylot: A modular platform for exploring latency-accuracy tradeoffs in autonomous vehicles},
  author={Gog, Ionel and Kalra, Sukrit and Schafhalter, Peter and Wright, Matthew A and Gonzalez, Joseph E and Stoica, Ion},
  booktitle={2021 IEEE International Conference on Robotics and Automation (ICRA)},
  pages={8806--8813},
  year={2021},
  organization={IEEE}
}

@inproceedings{shah2023gnm,
  title={GNM: A General Navigation Model to Drive Any Robot},
  author={Shah, Dhruv and Sridhar, Ajay and Bhorkar, Arjun and Hirose, Noriaki and Levine, Sergey},
  booktitle={2023 IEEE International Conference on Robotics and Automation (ICRA)},
  pages={7226--7233},
  year={2023},
  organization={IEEE}
}

@inproceedings{sridhar2024nomad,
  title={Nomad: Goal masked diffusion policies for navigation and exploration},
  author={Sridhar, Ajay and Shah, Dhruv and Glossop, Catherine and Levine, Sergey},
  booktitle={2024 IEEE International Conference on Robotics and Automation (ICRA)},
  pages={63--70},
  year={2024},
  organization={IEEE}
}

@article{cai2025navdp,
  title={NavDP: Learning Sim-to-Real Navigation Diffusion Policy with Privileged Information Guidance},
  author={Cai, Wenzhe and Peng, Jiaqi and Yang, Yuqiang and Zhang, Yujian and Wei, Meng and Wang, Hanqing and Chen, Yilun and Wang, Tai and Pang, Jiangmiao},
  journal={arXiv preprint arXiv:2505.08712},
  year={2025}
}

@article{xu2025navrl,
  title={Navrl: Learning safe flight in dynamic environments},
  author={Xu, Zhefan and Han, Xinming and Shen, Haoyu and Jin, Hanyu and Shimada, Kenji},
  journal={IEEE Robotics and Automation Letters},
  year={2025},
  publisher={IEEE}
}

@article{wang2025pi,
  title={$\pi^{3}$: Scalable Permutation-Equivariant Visual Geometry Learning},
  author={Wang, Yifan and Zhou, Jianjun and Zhu, Haoyi and Chang, Wenzheng and Zhou, Yang and Li, Zizun and Chen, Junyi and Pang, Jiangmiao and Shen, Chunhua and He, Tong},
  journal={arXiv preprint arXiv:2507.13347},
  year={2025}
}

@inproceedings{wang2025continuous,
  title={Continuous 3d perception model with persistent state},
  author={Wang, Qianqian and Zhang, Yifei and Holynski, Aleksander and Efros, Alexei A and Kanazawa, Angjoo},
  booktitle={Proceedings of the Computer Vision and Pattern Recognition Conference},
  pages={10510--10522},
  year={2025}
}

@inproceedings{yang2025fast3r,
  title={Fast3r: Towards 3d reconstruction of 1000+ images in one forward pass},
  author={Yang, Jianing and Sax, Alexander and Liang, Kevin J and Henaff, Mikael and Tang, Hao and Cao, Ang and Chai, Joyce and Meier, Franziska and Feiszli, Matt},
  booktitle={Proceedings of the Computer Vision and Pattern Recognition Conference},
  pages={21924--21935},
  year={2025}
}

@article{simeoni2025dinov3,
  title={Dinov3},
  author={Sim{\'e}oni, Oriane and Vo, Huy V and Seitzer, Maximilian and Baldassarre, Federico and Oquab, Maxime and Jose, Cijo and Khalidov, Vasil and Szafraniec, Marc and Yi, Seungeun and Ramamonjisoa, Micha{\"e}l and others},
  journal={arXiv preprint arXiv:2508.10104},
  year={2025}
}

@article{davison2007monoslam,
  title={MonoSLAM: Real-time single camera SLAM},
  author={Davison, Andrew J and Reid, Ian D and Molton, Nicholas D and Stasse, Olivier},
  journal={IEEE transactions on pattern analysis and machine intelligence},
  volume={29},
  number={6},
  pages={1052--1067},
  year={2007},
  publisher={IEEE}
}

@inproceedings{klein2009parallel,
  title={Parallel tracking and mapping on a camera phone},
  author={Klein, Georg and Murray, David},
  booktitle={2009 8th IEEE International Symposium on Mixed and Augmented Reality},
  pages={83--86},
  year={2009},
  organization={IEEE}
}

@article{engel2017direct,
  title={Direct sparse odometry},
  author={Engel, Jakob and Koltun, Vladlen and Cremers, Daniel},
  journal={IEEE transactions on pattern analysis and machine intelligence},
  volume={40},
  number={3},
  pages={611--625},
  year={2017},
  publisher={IEEE}
}

@inproceedings{engel2014lsd,
  title={LSD-SLAM: Large-scale direct monocular SLAM},
  author={Engel, Jakob and Sch{\"o}ps, Thomas and Cremers, Daniel},
  booktitle={European conference on computer vision},
  pages={834--849},
  year={2014},
  organization={Springer}
}

@inproceedings{godard2019digging,
  title={Digging into self-supervised monocular depth estimation},
  author={Godard, Cl{\'e}ment and Mac Aodha, Oisin and Firman, Michael and Brostow, Gabriel J},
  booktitle={Proceedings of the IEEE/CVF international conference on computer vision},
  pages={3828--3838},
  year={2019}
}

@article{mur2015orb,
  title={ORB-SLAM: A versatile and accurate monocular SLAM system},
  author={Mur-Artal, Raul and Montiel, Jose Maria Martinez and Tardos, Juan D},
  journal={IEEE transactions on robotics},
  volume={31},
  number={5},
  pages={1147--1163},
  year={2015},
  publisher={IEEE}
}

@inproceedings{kendall2015posenet,
  title={Posenet: A convolutional network for real-time 6-dof camera relocalization},
  author={Kendall, Alex and Grimes, Matthew and Cipolla, Roberto},
  booktitle={Proceedings of the IEEE international conference on computer vision},
  pages={2938--2946},
  year={2015}
}

@inproceedings{wimbauer2021monorec,
  title={MonoRec: Semi-supervised dense reconstruction in dynamic environments from a single moving camera},
  author={Wimbauer, Felix and Yang, Nan and Von Stumberg, Lukas and Zeller, Niclas and Cremers, Daniel},
  booktitle={Proceedings of the IEEE/CVF Conference on Computer Vision and Pattern Recognition},
  pages={6112--6122},
  year={2021}
}

@inproceedings{wei2024bev,
  title={Bev-odom: Reducing scale drift in monocular visual odometry with bev representation},
  author={Wei, Yufei and Lu, Sha and Han, Fuzhang and Xiong, Rong and Wang, Yue},
  booktitle={2024 IEEE/RSJ International Conference on Intelligent Robots and Systems (IROS)},
  pages={349--356},
  year={2024},
  organization={IEEE}
}

@article{shah2024codedvo,
  title={CodedVO: Coded Visual Odometry},
  author={Shah, Sachin and Rajyaguru, Naitri and Singh, Chahat Deep and Metzler, Christopher and Aloimonos, Yiannis},
  journal={IEEE Robotics and Automation Letters},
  year={2024},
  publisher={IEEE}
}

@article{qin2018vins,
  title={Vins-mono: A robust and versatile monocular visual-inertial state estimator},
  author={Qin, Tong and Li, Peiliang and Shen, Shaojie},
  journal={IEEE transactions on robotics},
  volume={34},
  number={4},
  pages={1004--1020},
  year={2018},
  publisher={IEEE}
}

@article{oquab2023dinov2,
  title={Dinov2: Learning robust visual features without supervision},
  author={Oquab, Maxime and Darcet, Timoth{\'e}e and Moutakanni, Th{\'e}o and Vo, Huy and Szafraniec, Marc and Khalidov, Vasil and Fernandez, Pierre and Haziza, Daniel and Massa, Francisco and El-Nouby, Alaaeldin and others},
  journal={arXiv preprint arXiv:2304.07193},
  year={2023}
}

@article{su2024roformer,
  title={Roformer: Enhanced transformer with rotary position embedding},
  author={Su, Jianlin and Ahmed, Murtadha and Lu, Yu and Pan, Shengfeng and Bo, Wen and Liu, Yunfeng},
  journal={Neurocomputing},
  volume={568},
  pages={127063},
  year={2024},
  publisher={Elsevier}
}

@article{ORBSLAM3_TRO,
  title={{ORB-SLAM3}: An Accurate Open-Source Library for Visual, Visual-Inertial and Multi-Map {SLAM}},
  author={Campos, Carlos and Elvira, Richard and G{\'o}mez, Juan J. and Montiel, Jos{\'e} M. M. and Tard{\'o}s, Juan D.},
  journal={IEEE Transactions on Robotics},
  volume={37},
  number={6},
  pages={1874--1890},
  year={2021}
}

@misc{interndata_n1,
  title={InternData-N1 Dataset},
  author={InternData-N1 Dataset contributors},
  howpublished={\url{https://huggingface.co/datasets/InternRobotics/InternData-N1}},
  year={2025},
  note         = {Accessed: 2025-09-15}
}

@inproceedings{grutopia,
    title={GRUtopia: Dream General Robots in a City at Scale},
    author={Wang, Hanqing and Chen, Jiahe and Huang, Wensi and Ben, Qingwei and Wang, Tai and Mi, Boyu and Huang, Tao and Zhao, Siheng and Chen, Yilun and Yang, Sizhe and Cao, Peizhou and Yu, Wenye and Ye, Zichao and Li, Jialun and Long, Junfeng and Wang, ZiRui and Wang, Huiling and Zhao, Ying and Tu, Zhongying and Qiao, Yu and Lin, Dahua and Pang Jiangmiao},
    year={2024},
    booktitle={arXiv},
}

@article{peng2025towards,
  title={Towards Latency-Aware 3D Streaming Perception for Autonomous Driving},
  author={Peng, Jiaqi and Wang, Tai and Pang, Jiangmiao and Shen, Yuan},
  journal={arXiv preprint arXiv:2504.19115},
  year={2025}
}

@article{wei2025ground,
  title={Ground Slow, Move Fast: A Dual-System Foundation Model for Generalizable Vision-and-Language Navigation},
  author={Wei, Meng and Wan, Chenyang and Peng, Jiaqi and Yu, Xiqian and Yang, Yuqiang and Feng, Delin and Cai, Wenzhe and Zhu, Chenming and Wang, Tai and Pang, Jiangmiao and others},
  journal={arXiv preprint arXiv:2512.08186},
  year={2025}
}
